\documentclass[11pt,a4paper]{article}
\usepackage[hyperref]{acl2017}
\usepackage{times}
\usepackage{latexsym}
\usepackage{graphicx}
\graphicspath{{images/}}
\usepackage{tabularx,booktabs}
\usepackage{makecell}
\usepackage{amssymb}
\usepackage{centernot}

\usepackage{url}

\aclfinalcopy % Uncomment this line for the final submission
\def\aclpaperid{41} %  Enter the acl Paper ID here

\title{Generating Steganographic Text with LSTMs}

\author{Tina Fang \\
  University of Waterloo \\
  {\small \tt tbfang@edu.uwaterloo.ca} \\\And
  Martin Jaggi \\
  EPFL \\
  {\small \tt martin.jaggi@epfl.ch} \\\And
  Katerina Argyraki \\
  EPFL \\
  {\small \tt katerina.argyraki@epfl.ch}
  }

\date{}

\begin{document}
\maketitle
\begin{abstract}

Motivated by concerns for user privacy, we design a steganographic system (``stegosystem'') that enables two users to exchange encrypted messages without an adversary detecting that such an exchange is taking place. We propose a new linguistic stegosystem based on a Long Short-Term Memory (LSTM) neural network. We demonstrate our approach on the Twitter and Enron email datasets and show that it yields high-quality steganographic text while significantly improving capacity (encrypted bits per word) relative to the state-of-the-art.
\end{abstract}

\section{Introduction}

The business model behind modern communication systems (email services 
or messaging services provided by social networks) is incompatible 
with end-to-end message encryption.
The providers of these services can afford to offer them free of charge
because most of their users agree to receive ``targeted ads''
(ads that are especially chosen to appeal to each user,
based on the needs the user has implied through their messages).
This model works as long as users communicate mostly in the clear,
which enables service providers to make informed guesses about user needs.

This situation does not prevent users from encrypting a few sensitive messages,
but it does take away some of the benefits of confidentiality.
For instance, imagine a scenario where two users want to exchange forbidden ideas
or organize forbidden events under an authoritarian regime;
in a world where most communication happens in the clear,
encrypting a small fraction of messages automatically makes these messages---and
the users who exchange them---suspicious.

With this motivation in mind, we want to design a system that enables two users to exchange encrypted messages, such that a passive adversary that reads the messages can determine neither the original content of the messages nor the fact that the messages are encrypted.

We build on linguistic steganography, 
i.e., the science of encoding a secret piece of information (``payload'') 
into a piece of text that looks like natural language (``stegotext''). 
We propose a novel stegosystem, based on a neural network, and demonstrate that it combines high quality of output 
(i.e., the stegotext indeed looks like natural language) 
with the highest capacity (number of bits encrypted per word) published in literature.

In the rest of the paper,
we describe existing linguistic stegosystems along with ours (\S\ref{steganography}),
provide details on our system (\S\ref{lstm}),
present preliminary experimental results on Twitter and email messages (\S\ref{experiments}),
and conclude with future directions (\S\ref{conclusion}).

\iffalse
Current communication systems (e.g. email services, Twitter, Facebook) conveniently allow open exchange of information between users. Yet, privacy concerns provide strong motivation to in some cases hide secret information (``payload") within non-secret information (``cover") in such public communication systems. Steganography is the art and science of using a method (``stegosystem") to hide payload with cover.
\fi

\section{Linguistic Steganography}
\label{steganography}

In this section, we summarize related work (\S\ref{related}),
then present out proposal (\S\ref{ours}).

\subsection{Related Work}
\label{related}
%\subsection{Cover Modification Systems}

Traditional linguistic stegosystems are based on modification of an existing cover text, 
e.g., using synonym substitution~\cite{topkara2006hiding,chang2014practical} and/or paraphrase substitution~\cite{chang2010linguistic}. 
The idea is to encode the secret information in the transformation of the cover text, 
ideally without affecting its meaning or grammatical correctness.
Of these systems, the most closely related to ours is CoverTweet~\cite{wilson2014linguistic}, 
a state-of-the-art cover modification stegosystem that uses Twitter as the medium of cover;
we compare to it in our preliminary evaluation (\S\ref{experiments}).

%\subsection{Alternative Stegosystems}

Cover modification can introduce syntactic and semantic unnaturalness~\cite{grosvald2011free};
to address this, Grovsald and Orgun proposed an alternative stegosystem where a human generates the stegotext manually, 
thus improving linguistic naturalness at the cost of human effort~\cite{grosvald2011free}.

Matryoshka~\cite{safaka2016matryoshka} takes this further:
in step 1, it generates candidate stegotext automatically based on an $n$-gram model of the English language;
in step 2, it presents the candidate stegotext to the human user for polishing, 
i.e., ideally small edits that improve linguistic naturalness.
However, the cost of human effort is still high,
because the (automatically generated) candidate stegotext is far from natural language,
and, as a result, the human user has to spend significant time and effort manually editing and augmenting it.

Volkhonskiy et al.\ have applied Generative Adversarial Networks \cite{goodfellow2014generative} to image steganography~\cite{volkhonskiy2017steganographic},
but we are not aware of any text stegosystem based on neural networks.

\subsection{Our Proposal: Steganographic LSTM}
\label{ours}

\begin{figure}[ht]
\includegraphics[width=\linewidth]{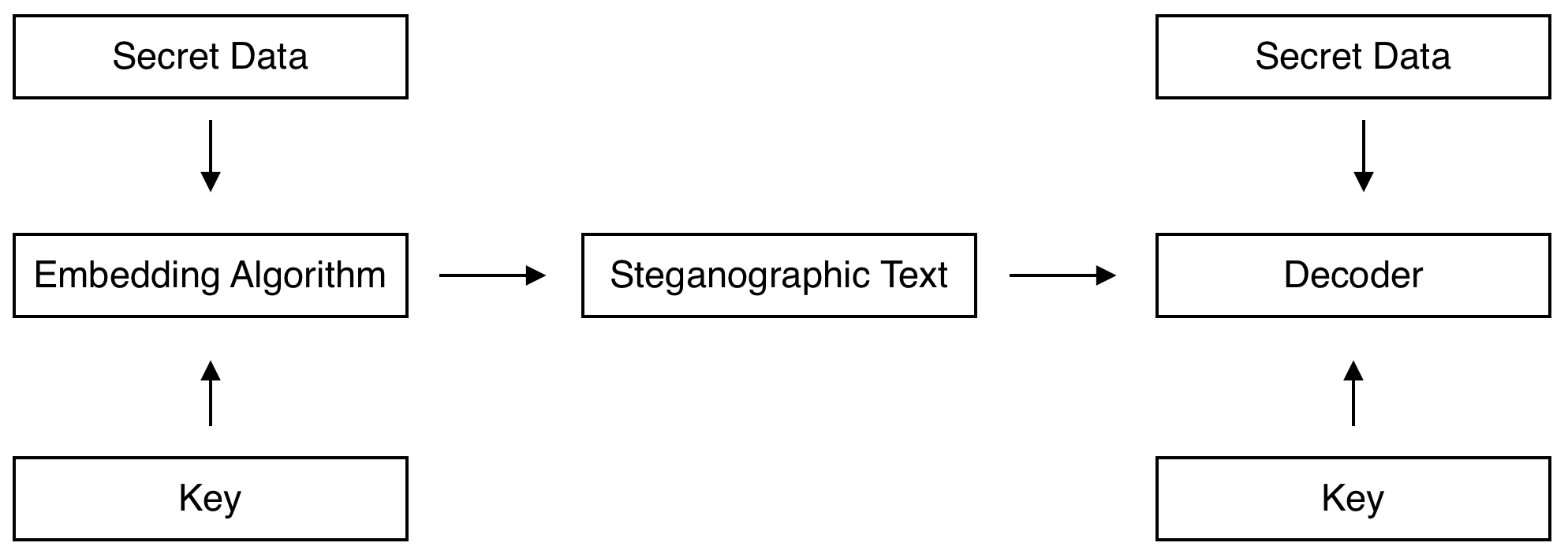}
\caption{\label{stegosystem-steps} Stegosystem building blocks.}
\vspace{-1mm}
\end{figure}

Motivated by the fact that LSTMs~\cite {hochreiter1997long} constitute the state of the art 
in text generation~\cite{jozefowicz2016exploring},
we propose to automatically generate the stegotext from an LSTM (as opposed to an $n$-gram model).
The output of the LSTM can then be used either directly as the stegotext,
or Matryoshka-style, i.e., as a candidate stegotext to be polished by a human user;
in this paper, we explore only the former option, i.e., we do not do any manual polishing.
We describe the main components of our system in the paragraphs below;
for reference, Fig.~\ref{stegosystem-steps} outlines the 
building blocks of a stegosystem \cite{salomon2003data}.

\paragraph{Secret data.}

The secret data is the information we want to hide. 
First, we compress and/or encrypt the secret data (e.g., in the simplest setting using the ASCII coding map) 
into a \emph{secret-containing bit string} $S$.
Second, we divide $S$ into smaller \emph{bit blocks} of length $|B|$, 
resulting in a total of $|S|/|B|$\footnote{If $|B| \centernot\mid |S|$, then we leave the remainder bit string out of encryption.} bit blocks. 
For example, if $S=100001$ and $|B|=2$, our bit-block sequence is $10$, $00$, $01$.
Based on this bit-block sequence, our steganographic LSTM generates words.

\paragraph{Key.} 

The sender and receiver share a key that maps bit blocks to token sets and is constructed as follows:
We start from the \emph{vocabulary}, which is the 
set of all possible tokens that may appear in the stegotext;
the tokens are typically words, but may also be punctuation marks. 
We partition the vocabulary into $2^{|B|}$ \emph{bins},
i.e., disjoint token sets, randomly selected from the vocabulary without replacement;
each token appears in exactly one bin, and each bin contains $|V|/2^{|B|}$ tokens. 
We map each bit block $B$ to a bin, denoted by $W_B$.
This mapping constitutes the shared key.
 
\begin{table}[ht]
\begin{center}
\begin{tabular}{|c|c|}
\hline \bf Bit Block & \bf Tokens \\ \hline
00 & This, am, weather, ... \\ \hline
01 & was, attaching, today, ... \\ \hline
10 & I, better, an, Great, ... \\ \hline
11 & great, than, NDA, ., ... \\
\hline
\end{tabular}
\end{center}
\caption{\label{bin-words} Example shared key.}
\end{table}

\paragraph{Embedding algorithm.}
\label{embedding-algo}

The embedding algorithm uses a modified word-level LSTM for language modeling~\cite{mikolov2010recurrent}.
To encode the secret-containing bit string $S$,
we consider one bit block $B$ at a time and have our LSTM select one token from bin $W_B$;
hence, the candidate stegotext has as many tokens as the number of bit blocks in $S$.
Even though we restrict the LSTM to select a token from a particular bin,
each bin should offer sufficient variety of tokens,
allowing the LSTM to generate text that looks natural.
For example, given the bit string ``1000011011'' and the key in Table~\ref{bin-words}, 
the LSTM can form the partial sentence in Table~\ref{word-gen}. 
We describe our LSTM model in more detail in the next section.

%we propose to base our embedding algorithm on LSTM models in order to generate such bin-restricted stegotexts.

\begin{table}[ht]
\begin{center}
\begin{tabular}{|c|ccccc|}
% \hline & \bf 1 & \bf 2 & \bf 3 & \bf 4 \\ \hline
\hline
\bf Bit String & 10 & 00 & 01 & 10 & 11 \\ \hline
\bf Token & I & am & attaching & an & NDA \\
\hline
\end{tabular}
\end{center}
\caption{\label{word-gen} Example stegotext generation.}
\end{table}

\paragraph{Decoder.}

The decoder recovers the original data deterministically and in a straightforward manner:
it takes as input the generated stegotext, considers one token at a time,
finds the token's bin in the shared key, and recovers the original bit block.

\paragraph{Common-token variant.}
We also explore a variant where we add a set of \emph{common tokens}, $C$, to all bins. 
These common tokens do not carry any secret information; they serve only to enhance stegotext naturalness.
When the LSTM selects a common token from a bin, 
we have it select an extra token from the same bin, until it selects a non-common token.
The decoder removes all common tokens before decoding.
We discuss the choice of common tokens and its implication on our system's performance in Section~\ref{experiments}.

\section{Steganographic LSTM Model}
\label{lstm}

In this section, we provide more details on our system:
how we modify the LSTM (\S\ref{modified}) and how we evaluate its output (\S\ref{metrics}).

\subsection{LSTM Modification}
\label{modified}

\paragraph{Text generation in classic LSTM.}

Classic LSTMs generate words as follows~\cite{sutskever2011generating}:
Given a word sequence $(x_1, x_2, \dots, x_T)$, the model has hidden states $(h_1, \dots, h_T)$, 
and resulting output vectors $(o_1,\dots,o_T)$. 
Each output vector $o_t$ has length $|V|$, 
and each output-vector element $o_t^{(j)}$ is the unnormalized probability of word $j$ in the vocabulary. 
Normalized probabilities for each candidate word are obtained by the following softmax activation function:
$$\mathop{softmax}(o_t)_j := \exp(o_t^{(j)})\Big/\sum_{k} \exp(o_t^{(k)}).$$ 
The LSTM then selects the word with the highest probability $P[ x_{t+1}\,|\,x_{\leq t}]$ as its next word.

\paragraph{Text generation in our LSTM.}

In our steganographic LSTM, word selection is restricted by the shared key.
That is, given bit block $B$, the LSTM has to select its next word from bin~$W_B$. 
We set $P[x = w_j] = 0$ for $j \notin W_B$, so that the multinomial softmax function selects 
the word with the highest probability within~$W_B$.

\paragraph{Common tokens.}

In the common-token variant, we restrict $P[x = w_j] = 0$ only for $j \notin (W_B \cup C)$, 
where $C$ is the set of common tokens added to all bins.

\subsection{Evaluation Metrics}
\label{metrics}

We use perplexity to quantify stegotext quality;
and capacity (i.e., encrypted bits per output word) to quantify its efficiency
in carrying secret information.
In Section \ref{experiments},
we also discuss our stegotext quality as empirically perceived by us as human readers.

\paragraph{Perplexity.}

Perplexity is a standard metric for the quality of language models~\cite{martin2000speech}, 
and it is defined as the average per-word log-probability on the valid data set: 
$\exp(-1/N \sum_{i} \ln p[w_i] )$ \cite{jozefowicz2016exploring}. 
Lower perplexity indicates a better model. 

In our steganographic LSTM, we cannot use this metric as is:
since we enforce $p[w_i] = 0$ for $w_i \notin W_B$, 
the corresponding $\ln p[w_i]$ becomes undefined under this vocabulary.

Instead, we measure the probability of $w_i$ by taking the average of $p[w_i]$ over all possible secret bit blocks $B$,
under the assumption that bit blocks are distributed uniformly. 
By the Law of Large Numbers \cite{revesz2014laws},
if we perform many stegotext-generating trials using different random secret data as input, 
the probability of each word will tend to the expected value, $\sum p[w_{i},B] / 2^{|B|}$, 
Hence, we set $p[w_i] := \sum p[w_{i},B] / 2^{|B|}$ instead of $p[w_i] = 0$ for $w_i \notin W_B$.

\paragraph{Capacity.}

Our system's capacity is the number of encrypted bits per output word.
Without common tokens, capacity is always $|B|$ bits/word
(since each bit block of size $|B|$ is always mapped to one output word).
In the common-token variant, 
capacity decreases because the output includes common tokens that do not carry any secret information;
in particular, if the fraction of common tokens is $p$, 
then capacity is $(1-p) \cdot |B|$.

\section{Experiments}
\label{experiments}

In this section, we present our preliminary experimental evaluation:
our Twitter and email datasets (\S\ref{datasets}),
details about the LSTMs used to produce our results (\S\ref{impl}),
and finally a discussion of our results (\S\ref{results}).

\subsection{Datasets}
\label{datasets}

Tweets and emails are among the most popular media of open communication 
and therefore provide very realistic environments for hiding information. 
We thus trained our LSTMs on those two domains, Twitter messages and Enron emails \cite{klimt2004enron}, which vary greatly in message length and vocabulary size.

For Twitter, we used the NLTK tokenizer to tokenize tweets \cite{bird2006nltk} into words and punctuation marks. 
We normalized the content by replacing usernames and URLs with a username token ($<$user$>$) and a URL token ($<$url$>$), 
respectively. 
We used $600$ thousand tweets with a total of $45$ million words and a vocabulary of size $225$ thousand.

For Enron, we cleaned and extracted email message bodies \cite{zhou2007strategies} from the Enron dataset, 
and we tokenized the messages into words and punctuation marks. 
We took the first $100$MB of the resulting messages, with $16.8$ million tokens and a vocabulary size of $406$ thousand.

\subsection{Implementation Details}
\label{impl}

We implemented multi-layered LSTMs based on PyTorch\footnote{https://github.com/pytorch} in both experiments. We did not use pre-trained word embeddings \cite{mikolov2013distributed,pennington2014glove}, and instead trained word embeddings of dimension $200$ from scratch. 

We optimized with Stochastic Gradient Descent and used a batch size of $20$. The initial learning rate was $20$ and the decay factor per epoch was~$4$. The learning rate decay occurred only when the validation loss did not improve. Model training was done on an NVIDIA GeForce GTX TITAN~X.

For Twitter, we used a $2$-layer LSTM with $600$ units, unrolled for $25$ steps for back propagation. We clipped the norm of the gradients \cite{pascanu2013difficulty} at $0.25$ and applied $20$\% dropout \cite{srivastava2014dropout}. We stopped the training after $12$ epochs ($10$ hours) based on validation loss convergence.

For Enron, we used a $3$-layer LSTM with $600$ units and no regularization. We unrolled the network for $20$ steps for back propagation. We stopped the training after $6$ epochs ($2$ days).

\subsection{Results and Discussion}
\label{results}

\subsubsection{Tweets}
We evaluate resulting tweets generated by LSTMs of $1$ (non-steganographic), $2$, $4$, $8$ bins. Furthermore, we found empirically that adding $10$ most frequent tokens from the Twitter corpus was enough to significantly improve the grammatical correctness and semantic reasonableness of tweets. Table \ref{tweet-ppl} shows the relationship between capacity (\emph{bits per word}), and quantitative text quality (\emph{perplexity}). It also compares models with and without adding common tokens using perplexity and bits per word.

Table \ref{common-comparison} shows example output texts of LSTMs with and without common tokens added. To reflect the variation in the quality of the tweets, we represent tweets that are good and poor in quality\footnote{For each category, we manually evaluate ~$60$ randomly generated tweets based grammatical correctness, semantic coherence, and resemblance to real tweets. We select tweets from the $25$th, $50$th, and $75$th percentile, and call them ``good", ``average", and ``poor" respectively. We limit to tweets that are not offensive in language.}. 

We replaced $<$user$>$  generated by the LSTM with mock usernames for a more realistic presentation in Table \ref{common-comparison}. In practice, we can replace the $<$user$>$ tokens systematically, randomly selecting followers or followees of that tweet sender, for example. 

Re-tweet messages starting with ``RT" can also be problematic, because it will be easy to check whether the original message of the retweeted message exists. A simple approach to deal with this is to eliminate ``RT" messages from training (or at generation). Finally, since we made all tweets lower case in the pre-processing step, we can also post-process tweets to adhere to proper English capitalization rules.

%We post-processed the tweets by un-tokenization of punctuations. 

\begin{table}[ht]
\begin{center}
{\small
\begin{tabular}{|c|c|c|c|c|}
\hline & \multicolumn{2}{c|}{\textbf{Original}} & \multicolumn{2}{c|}{\textbf{Common Tokens}} \\
\hline \bf \# of Bins & bpw &  ppl &  bpw & ppl \\ \hline
\textbf{$1$} & $0$  & $134.73$ & $0$ & $134.73$ \\ \hline
\textbf{$2$} & $1$ & $190.84$ & $0.65$ & $171.35$ \\ \hline
\textbf{$4$} & $2$ & $381.2$ & $1.17$ & $277.55$ \\ \hline 
\textbf{$8$} & $3$ & $833.11$ & $1.53$ & $476.66$ \\ \hline
\end{tabular}
}
\end{center}
\caption{\label{tweet-ppl} An increase of of capacity correlates with an increase of perplexity, which implies that there is a negative correlation between capacity and text quality. After adding common tokens, there is a significant reduction in perplexity \emph{(ppl)}, at the expense of a lower capacity \emph{(bits per word)}.}
\end{table}

\begin{table*}[ht]
\begin{center}
{\small
\begin{tabularx}{\linewidth}{|c|X|X| }
\hline \bf \# of Bins & \bf Tweets & \bf Tweets with Common Tokens \\ \hline
\textbf{$2$} & \begin{minipage}{\linewidth}\textbf{good:} i was just looking for someone that i used have.\\
\textbf{poor:} cry and speak! rt @user421: relatable personal hygiene for body and making bad things as a best friend in lifee \end{minipage}
 & \begin{minipage}{\linewidth}
\textbf{good:} i'm happy with you. i'll take a pic   \\
\textbf{poor:} rt: cut your hair, the smallest things get to the body. 
\end{minipage} \\ \hline

\textbf{$4$} & \begin{minipage}{\linewidth}
\textbf{good:} @user390 looool yeah she likes me then ;). you did? \\
\textbf{poor:} ``where else were u making?... i feel fine? - e? lol" * does a voice for me \& take it to walmart? \end{minipage}
& \begin{minipage}{\linewidth}
\textbf{good:} i just wanna move. collapses. \\
\textbf{poor:} i hate being contemplating for something i want to.  
\end{minipage} \\ \hline

\textbf{$8$} &\begin{minipage}{\linewidth}
\textbf{good:} @user239 hahah. sorry that my bf is amazing because i'm a bad influence ;). \\
\textbf{poor:} so happy this to have been working my ass and they already took the perfect. but it's just cause you're too busy the slows out! love... * dancing on her face, holding two count out cold * ( a link with a roof on punishment... - please :) \end{minipage}
 & \begin{minipage}{\linewidth}
\textbf{good:} i hate the smell of my house. \\
\textbf{poor:} a few simple i can't. i need to make my specs jump surprisingly. 
\end{minipage} \\ \hline
\end{tabularx}
}
\end{center}
\caption{\label{common-comparison} We observe that the model with common tokens produces tweets simpler in style, and uses more words from the set of common tokens. There is a large improvement in grammatical correctness and context coherence after adding common tokens, especially in the ``poor" examples. For example, adding the line break token %, ``\textless br /\textgreater" 
reduced the length of the tweet generated from the 8-bin LSTM.
}
\end{table*} 

\subsubsection{Emails}

We also tested email generation, and Table \ref{email-bin} shows sample email passages\footnote{We only present passages "average" in quality to conserve space.} from each bin. We post-processed the emails with untokenization of punctuations. 

The biggest difference between emails and tweets is that emails have a much longer range for context dependency, with context spanning sentences and paragraphs. This is challenging to model even for the non-steganographic LSTM. Once the long-range context dependency of the non-steganographic LSTM improves, the context dependency of the steganographic LSTMs should also improve.

\begin{table}[ht]
\begin{center}
{\small
\begin{tabularx}{\linewidth}{ |c | X| }
\hline \bf \# of Bins & \bf Sample Email \\ \hline
\textbf{$1$} & -----Original Message----- From: Nelson, Michelle Sent: Thursday, January 03, 2002 3:35 PM To: Maggi, Mike Subject: Hey, You are probably a list of people that are around asleep about the point of them and our wife. Rob \\ \hline
\textbf{$2$} & If you do like to comment on the above you will not contact me at the above time by 8:00 a.m. on Monday, March 13 and July 16 and Tuesday, May 13 and Tuesday, March 9 - Thursday, June 17, - 9:00 to 11:30 AM. \\ \hline
\textbf{$4$} & At a moment when my group was working for a few weeks, we were able to get more flexibility through in order that we would not be willing. \\ \hline
\end{tabularx}
}
\end{center}
\caption{\label{email-bin} The issue of context inconsistency is present for all bins. However, the resulting text remains syntactical even as the number of bins increases.}
\end{table}

\subsection{Comparison with Other Stegosystems}
\label{comparisons}

For all comparisons, we use our 4-bin model with no common tokens added. 

Our model significantly improves the state-of-the-art capacity. Cover modification based stegosystems hide $1$-$2$ bits per sentence \cite{chang2012adjective}. The state-of-the-art Twitter stegosystem hides $2.8$ bits of per tweet \cite{wilson2016avoiding}. Assuming $16.04$ words per tweet\footnote{Based a random sample of $2$ million tweets.}, our $4$-bin system hides \textbf{32 bits per tweet}, over $11$ times higher than \cite{wilson2016avoiding}.

We hypothesize that the subjective quality of our generated tweets will be comparable to tweets produced by CoverTweet \shortcite{wilson2014linguistic}. We present some examples\footnote{Tweets selected for comparison are ``average" in quality.} in Table \ref{tweet-quality} to show there is potential for a comparison. This contrasts the previous conception that cover generation methods are fatally weak against human judges \cite{wilson2014linguistic}. CoverTweet was tested to be secure against human judges. Formal experiments will be necessary to establish that our system is also secure against human judges.

\begin{table}[ht]
\begin{center}
{\small
\begin{tabularx}{\linewidth}{|X|X| }
\hline \bf CoverTweet \shortcite{wilson2014linguistic} & \bf Steganographic LSTM \\ \hline
yall must have 11:11 set 1 minute early before yall tweet it, because soon as 11:11 hit yall don't wastes no time. lol & i wanna go to sleep in the gym, ny in peoples houses \& i'm in the gym..! :(( \\ \hline
you can tell when somebody hating on you! & i would rather marry a regular sunday!! \\ \hline
most of the people who got mouth can't beat you. & my mom is going so hard to get his jam. \\ \hline
\end{tabularx}
}
\end{center}
\caption{\label{tweet-quality} The tweets generated by the 4-bin LSTM ($32$ bits per tweet) are reasonably comparable in quality to tweets produced by CoverTweet ($2.8$ bits per tweet).}
\end{table} 

Our system also offers flexibility for the user to freely trade-off capacity and text quality. Though we chose the 4-bin model with no common tokens for comparison, user can choose to use more bins to achieve an even higher capacity, or use less bins and add common tokens to increase text quality. This is not the case with existing cover modification systems, where capacity is bounded above by the number of transformation options \cite{wilson2014linguistic}.

\section{Conclusion and Future Work}
\label{conclusion}

In this paper, we opened a new application of LSTMs, namely, steganographic text generation. We presented our steganographic model based on existing language modeling LSTMs, and demonstrated that our model produces realistic tweets and emails while hiding information. 

In comparison to the state-of-the-art steganographic systems, our system has the advantage of encoding much more information (around $2$ bits per word). This advantage makes the system more usable and scalable in practice.

In future work, we will formally evaluate our system's security against human judges and other steganography detection (steganalysis) methods \cite{wilson2015detection,kodovsky2012ensemble}.
When evaluated against an automated classifier, the setup becomes that of a Generative Adversarial Network \cite{goodfellow2014generative}, though with additional conditions for the generator (the secret bits) which are unknown to the discriminator, and not necessarily employing joint training. 
Another line of future research is to generate tweets which are personalized to a user type or interest group, instead of reflecting all twitter users. Furthermore,  we plan to explore if capacity can be improved even more by using probabilistic encoders/decoders, as e.g. in Matryoshka \cite[Section~4]{safaka2016matryoshka}.

Ultimately, we aim to open-source our stegosystem so that users of open communication systems (e.g. Twitter, emails) can use our stegosystem to communicate private and sensitive information.

\clearpage
\bibliography{references}
\bibliographystyle{acl_natbib}

% \appendix

\end{document}